\documentclass[runningheads]{llncs}

\usepackage[mobile]{eccv}

\usepackage{eccvabbrv}
\usepackage{adjustbox}

\usepackage{graphicx}
\usepackage{booktabs}

\usepackage{hyperref}

\begin{document}

\title{Compact 3D Scene Representation via Self-Organizing Gaussian Grids}

\titlerunning{Abbreviated paper title}

\author{
Wieland Morgenstern\inst{1} \and Florian Barthel\inst{1,2} \and Anna Hilsmann\inst{1} \and \\
Peter Eisert\inst{1,2}
}

\authorrunning{W. Morgenstern et al.}

\institute{Fraunhofer Heinrich Hertz Institute, HHI \\
\and
Humboldt University of Berlin \\
\email{\{first\}.\{last\}@hhi.fraunhofer.de}\\
}

\maketitle

\begin{abstract} 
3D Gaussian Splatting has recently emerged as a highly promising technique for modeling of static 3D scenes. In contrast to Neural Radiance Fields, it utilizes efficient rasterization allowing for very fast rendering at high-quality. However, the storage size is significantly higher, which hinders practical deployment, e.g.~on resource constrained devices. In this paper, we introduce a compact scene representation organizing the parameters of 3D Gaussian Splatting (3DGS) into a 2D grid with local homogeneity,  ensuring a drastic reduction in storage requirements without compromising visual quality during rendering.
Central to our idea is the explicit exploitation of perceptual redundancies present in natural scenes. In essence, the inherent nature of a scene allows for numerous  permutations of Gaussian parameters to equivalently represent it. To this end, we propose a novel highly parallel algorithm that regularly arranges the high-dimensional Gaussian parameters into a 2D grid while preserving their neighborhood structure. During training, we further enforce local smoothness between the sorted parameters in the grid. The uncompressed Gaussians use the same structure as 3DGS, ensuring a seamless integration with established renderers. Our method achieves a reduction factor of 17x to 42x in size for complex scenes with no increase in training time, marking a substantial leap forward in the domain of 3D scene distribution and consumption.
Additional information can be found on our project page: \href{https://fraunhoferhhi.github.io/Self-Organizing-Gaussians/}{fraunhoferhhi.github.io/Self-Organizing-Gaussians/}

\end{abstract}

\begin{figure}[t]
  \centering
   \includegraphics[width=0.8\linewidth]{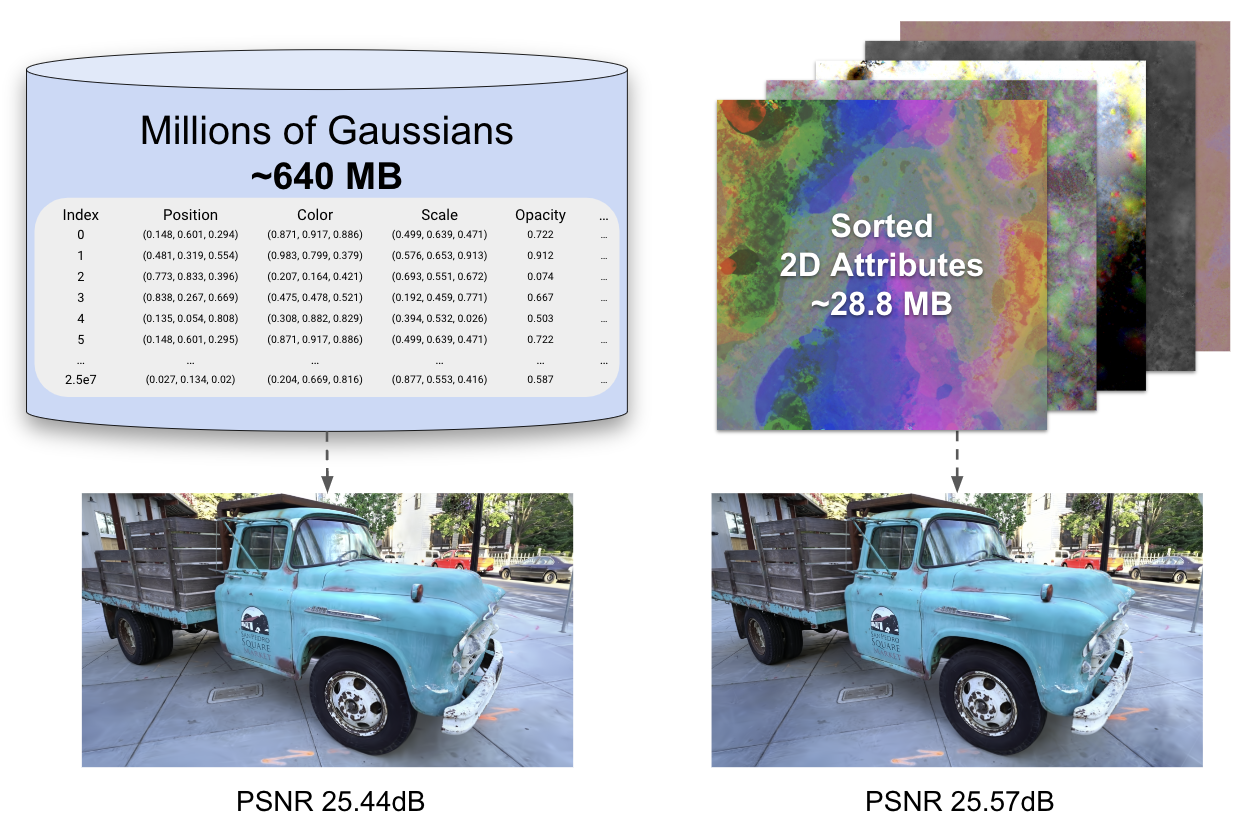}
   \vspace{-0.2cm}
   \caption{Our 3DGS training method allows for high compression of 3DGS attributes while maintaining high rendering quality. By sorting the Gaussian features periodically into a 2D grid and applying a smoothness loss during scene generation, we significantly reduce the storage footprint using state-of-the-art image compression methods.}
   \label{fig:teaser}
\end{figure}

\section{Introduction}
\label{sec:intro}

3D reconstruction and rendering of real-world scenes has been a cornerstone of computer vision since its inception. Neural Radiance Fields (NeRFs) \cite{mildenhall2020} and 3D Gaussian Splatting (3DGS) \cite{kerbl2023} have recently revolutionized 3D representation and novel-view synthesis.
Assessing the efficacy of 3D scene reconstruction methods and representations typically involves three important metrics: rendering quality, rendering speed, and storage size. While rendering quality is of general importance, rendering speed and storage size play a crucial role especially for real time applications, portability on devices with limited performance and memory or, for fast web applications.

NeRF-based rendering methods achieve high rendering quality \cite{barron2021}.  However, they are costly to train and synthesis speed is slow, as each pixel value is computed by aggregating multiple results from a forward pass through a neural network. %
Recent methods have enabled considerably faster training and inference \cite{fridovich2022,sun2022,reiser2023}, often, however, at the cost of decreased quality and/or increased disk memory for each model. A recent alternative to NeRF is 3D Gaussian Splatting (3DGS), explicitly representing the scene with localized 3D Gaussian splats \cite{kerbl2023}. In contrast to NeRFs, 3DGS employs an efficient rasterization algorithm allowing for very fast rendering at high quality. In addition, the representation enables the modeling of consistent object motion in dynamic scenes \cite{luiten2023dynamic}. On the downside, however, the storage size of 3DGS models is significantly higher than with NeRFs, because 3DGS stores millions of parameters in a large unorganized list. Such a data structure can take up to several hundred megabytes depending on the scene, which hinders practical deployment, e.g. on resource constrained devices.

To address this limitation, we propose a new compact representation based on 3DGS, allowing for efficient encoding and storage. We exploit the fact that numerous permutations of Gaussian parameters can equivalently represent a scene with the same visual quality and arrange the millions of multidimensional list entries of 3DGS into a structured 2D grid with increased smoothness for efficientencoding.

In addition, we integrate a smoothness loss into the 3DGS scene optimization process to guide the ambiguous arrangements of the splats towards a visually equivalent but better compressible parameter set. 
We finally quantize the parameters and encode them by efficient 2D image compression methods, reducing the overall file size by a factor of 17x to 42x, without sacrificing rendering quality.

With our method, we enable novel-view synthesis at high quality, high rendering speed, and with a small storage footprint, bringing high quality 3D scene reconstruction and rendering one step closer towards applications on small devices with limited storage capacity or fast web applications. 

Our main contributions are:

\begin{enumerate}
    \item We propose a new compact scene representation and training concept for 3DGS, structuring the high-dimensional features in a smooth 2D grid, which can be efficiently encoded using state-of-the-art compression methods.
    \item We introduce an efficient 2D sorting algorithm called \textit{Parallel Linear Assignment Sorting (PLAS)} that sorts millions of 3DGS parameters on the GPU in seconds.
    \item We provide a simple to use interface for compressing and decompressing the resulting 3D scenes. The decompressed reconstructions share the structure of 3DGS, allowing integration into established renderers.
    \item We efficiently reduce the storage size by a factor of 17x to 42x while maintaining high visual quality

\end{enumerate}

\section{State of the Art}
\label{sec:sota}
We first overview recent advances in 3D reconstruction and representation, before we take a closer look at high dimensional data structuring and sorting.

\subsection{3D Scene Representation}
Over the past years, 3D scene representation has seen rapid improvements, especially with the advent of deep learning. The general goal is to allow novel-view synthesis of 3D objects captured by a number of 2D images. In the following, we will give a brief overview of the most relevant technologies and highlight their key differences. For a more comprehensive review in this field, please refer to Tewari \etal \cite{tewari2022}.

Early novel-view synthesis approaches were based on pure image interpolation as for the lightfields \cite{gortler1996, levoy1996}, or reconstructed 3D geometry by means of regular voxel grids \cite{seitz1999}, point clouds, or meshes. Structure-from-Motion (SfM) and Multi-View-Stereo (MVS) were proposed to estimate such representations from a set of input images \cite{dellaert2000, snavely2006, goesele2007}. These methods were the basis for view-synthesis approaches guided by explicit representations of the geometry \cite{chaurasia2013, kopanas2021, rueckert2022}. Also, compression of the large 3D point cloud datasets has been addressed \cite{graziosi2020,quach2022}, but is typically restricted to simple colored points without additional attributes.

In recent years, implicit representations have been proposed to represent geometry \cite{sitzmann2019}. Especially, Neural Radiance Fields (NeRFs) have revolutionized novel-view synthesis \cite{mildenhall2020} by representing a scene volume as a multi-layer perceptron (MLP). This representation allows the visualization of thin structures, semi-transparent surfaces and highly realistic light reflections. In general, NeRFs show an outstanding rendering quality, however, they are slow during training and inference.

The success of NeRF has resulted in numerous follow-up methods improving rendering quality, e.g.~for large scenes \cite{martin-brualla2021}, or speed \cite{barron2021, mueller2022, fridovich2022, takikawa2022}. While Mip-NeRF360 \cite{barron2021} as one of the state-of-the-art approaches achieves an outstanding quality, it is still very costly to train and render. Methods focusing on faster training and/or rendering exploit spatial data structures and encodings in order to accelerate computation, representing the scene as a grid \cite{fridovich2022}, tensors \cite{chen2022}, latent features \cite{liu2020,karnewar2022,sun2022}, hash grids \cite{mueller2022} or using vector quantizations \cite{ takikawa2022}. Among these, InstantNGP is a widely recognized example, using a smaller MLP to represent density and appearance though the use of a hash table and an occupancy grid \cite{mueller2022}. Plenoxels estimate explicit spherical harmonics in addition to color and density to represent view dependent appearance effects \cite{fridovich2022}. In order to improve the rendering, the values are interpolated from a sparse 3D grid of features using trilinear interpolation, resulting in considerably higher rendering speed at the cost of lower quality. In addition, the storage size of Plenoxels is high, given that the number of elements in the 3D grid grows by $\mathcal{O}(n^3)$. 

Other approaches aiming at real-time rendering involved precomputing and storing NeRF's view-dependent colors and opacities in volumetric data structures \cite{garbin2021,hedman2021,yu2021,chen2022,zhang2022,lee2023} or partitioning the scene into voxels, each represented by separate small MLPs. However, these representations suffer from quality degradation and impose substantial graphics memory demands, restricting their application to objects rather than entire scenes. Several methods have been proposed to reduce the required disk space. Among these, Memory-efficient NeRF (MERF) \cite{lee2023} and TensoRF \cite{chen2022}, share a similar approach, storing the 3D scene representation inside lightweight 2D features. Other methods to compress NeRF reconstructions use incremental pruning \cite{deng2023}, vector quantization \cite{lee2023, zhong2023} or compressed codebook representations \cite{takikawa2022}.  However, these approaches still require costly network inference passes for each query point in the scene and are therefore not capable of real-time rendering of large scenes. 

Our approach is inspired by a recently proposed approach for 3D scene representation by Kerbl \etal~\cite{kerbl2023} combining the advantages of explicit representations (fast training and rendering) with those of implicit representation (stochastic sampling) by representing the scene by millions of 3D Gaussians, called 3D Gaussian Splatting (3DGS). The parameters (3D coordinates, color, size, density, orientation, and spherical harmonics) of these Gaussians are optimized
using an efficient differentiable rasterization algorithm starting from a set of Structure-from-Motion points to best represent a given 3D scene. The main contribution of this approach is a representation that is continuous and differentiable while also allowing fast rendering: 3DGS is currently the state-of-the-art approach in terms of quality, rendering time, and performance. However, this comes at the cost of significantly higher storage requirements, as for each scene, millions of high dimensional Gaussians are stored in an unstructured list on disk.

\subsection{Mapping Higher Dimensional Data to 2D Grids}
\label{sec:sota_mapping}

The ambiguity of ordering of the high dimensional point clouds in approaches like 3DGS \cite{kerbl2023} enables optimization of point indices for more efficient representation and storage.
Grid-based sorting techniques like Self-Organizing Map \cite{Kohonen1982, Kohonen2013}, Self-Sorting Map \cite{Strong2013, Strong2014}, or Fast Linear Assignment Sorting \cite{barthel2023}, allow mapping high-dimensional data points onto a 2D grid, organized by similarity, thereby facilitating an efficient compression of the data.

A self-organizing map (SOM) \cite{Kohonen1982, Kohonen2013} is a neural network trained using unsupervised learning to produce a low-dimensional (typically two-dimensional), discretized representation of the input space of the training samples.
Due to the sequential nature of the SOM training algorithm, the last input vectors can only be assigned to the few remaining unassigned map positions, resulting in isolated and poorly positioned vectors within the map. Self-Sorting Maps \cite{Strong2013, Strong2014} avoid this problem by swapping assigned positions of four input vectors at a time. Due to the factorial number of permutations, adding more candidates would be computationally too complex. Linear Assignment Sorting (LAS) \cite{barthel2023} outperforms other sorting methods in terms of speed and sorting quality for non-parallel computations, by swapping all (or several (Fast LAS)) vectors simultaneously using a continuously filtered map. %

However, these grid-sorting algorithms, effective for handling thousands of items, prove inadequate for our specific application, where we seek to structure millions of Gaussians spread across various dimensions. To overcome this, we adapt and expand upon existing sorting methodologies, enabling continuous sorting of Gaussian data during training without inflating the training duration.

Our goal is a new representation and training scheme for 3D Gaussian Splatting with drastically reduced storage requirements while maintaining visual quality during rendering. 
The idea of our method is to map originally unstructured 3D Gaussians to a 2D grid in such a way that local spatial relationships are preserved. By mapping all Gaussian parameters (position, color, etc.) to the same 2D position, we effectively create multiple data layers with the same layout. In this 2-dimensional layout, neighboring Gaussians will have similar attribute values, facilitating compression and enabling the storage of the same scene with much less disk space.%

\begin{figure*}[ht]
  \centering
   \includegraphics[width=1.0\linewidth]{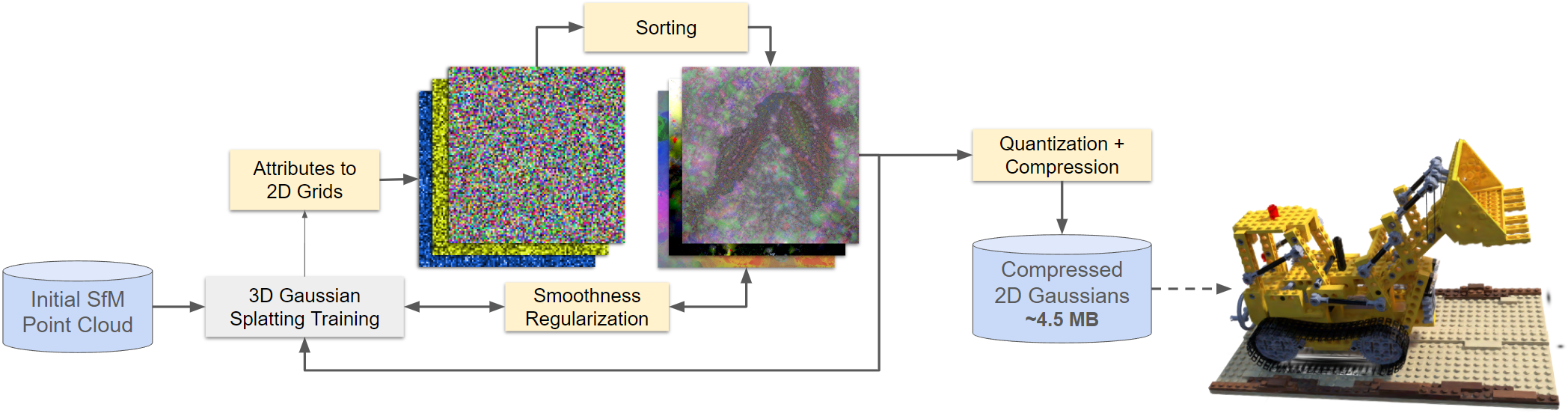}
   \vspace{-0.2cm}
   \caption{An overview of our novel 3DGS training method. During training, we arrange all high dimensional attributes into multiple 2D grids. Those grids are sorted and a smoothness regularization is applied. This creates redundancy which help to compress the 2D grids into small files using off-the-shelf compression methods.}
   \label{fig:method}
\end{figure*}

\section{Method}
\label{sec:method}
\begin{figure*}[htbp]
  \centering
   \includegraphics[width=0.75\linewidth]{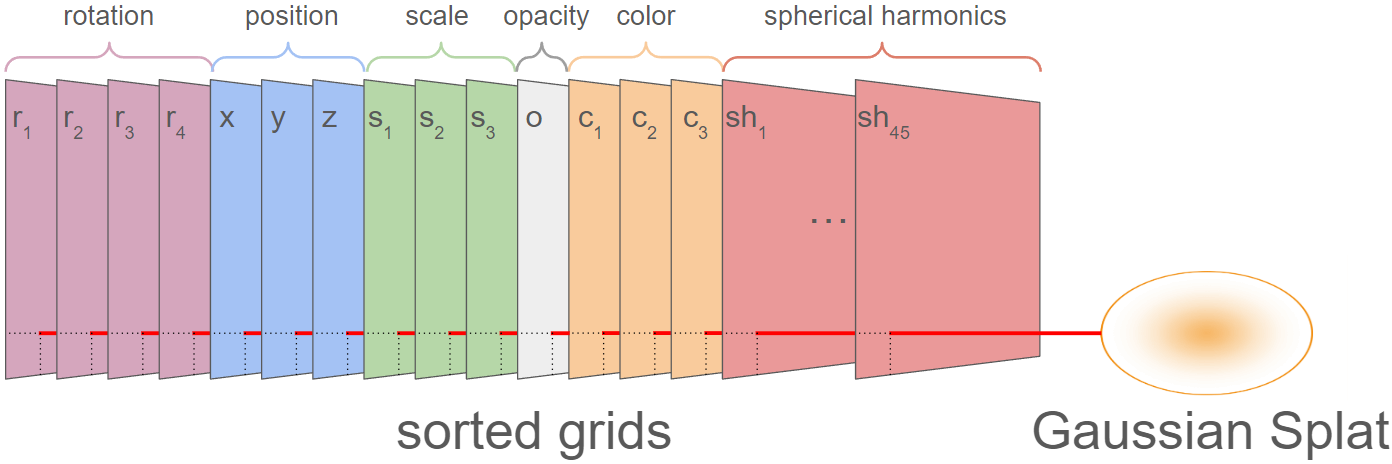}
   \vspace{-0.2cm}
   \caption{Visualization of the conversion of 2D attribute grids into Gaussian splat objects. Note that the sorted grids have to align. Therefore, they cannot be sorted individually. Here, color represents the DC components of the spherical harmonics.}
   \label{fig:grid2splat}
\end{figure*}

Figure \ref{fig:method} illustrates an outline of our method. Our training introduces two new components to the 3DGS framework: a highly-parallel 2D sorting step (Section \ref{subsec:sort}), and an enhanced scene optimization with a local neighborhood loss on the 2D grid (Section \ref{subsec:smooth}). The sorting step initializes the optimization process by explicitly organizing the values, establishing an initial smooth local neighborhood without requiring a global optimization. 
Once sorted, our enhanced training incorporates a neighborhood loss, further enforcing local smoothness on the 2D grid, while the differentiable renderer of the 3D Gaussian splatting optimizes the Gaussians attributes to accurately represent the scene. 
Similar to the original 3DGS approach, Gaussians are pruned from the scene at regular intervals, and new Gaussians are added in under-reconstructed parts of the scene. This step is called densification. After each densification step, we re-sort all Gaussians into the 2D grid. This step re-establishes local neighborhoods after pruning and densification may have disrupted them. The attribute channels cannot be sorted individually, as each position on the grid belongs to the same Gaussian splat object, as visualized in figure \ref{fig:grid2splat}. Therefore, the sorting algorithm has to find one permutation that satisfies all attributes at once. Otherwise, it would require additional storage size to align the attributes of the sorted grids.

To store the Gaussian attributes, we leverage off-the-shelf compression methods (described in Section\ref{sec:compression}), which can be applied effectively to the well-organized data.

\begin{figure*}[t]
  \centering
   \includegraphics[width=\linewidth]{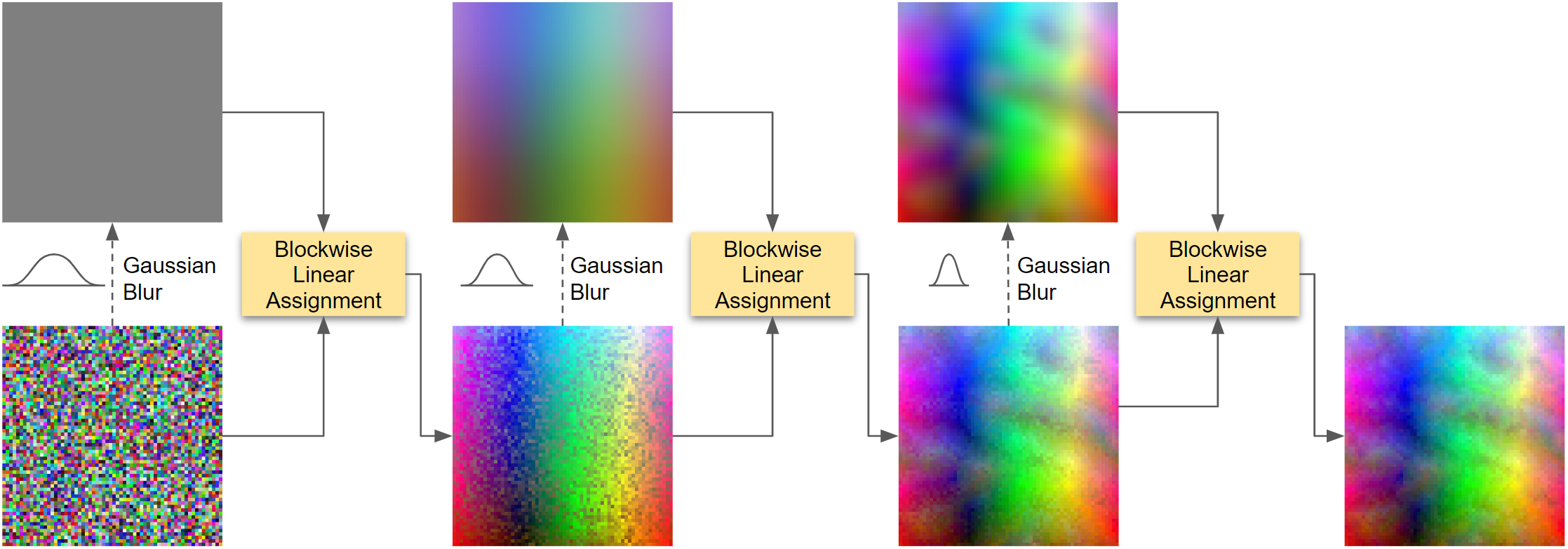}
   \vspace{-0.2cm}
   \caption{We sort the Gaussians into a 2D grid with an iterative approach. Gaussians are assigned to the position matching closest to a smoothed version of the grid. With increasing iterations, we decrease the kernel size and sigma of the 2D gaussian filter.}
   \label{fig:blockwise_assignment}
\end{figure*}

\subsection{Sorting High Dimensional Gaussians into 2D Grids}
\label{subsec:sort}

Traditional grid-sorting algorithms are suitable for thousands of items, yet fall short for our application, where even small scenes contain millions of Gaussians across multiple dimensions. To address this, we combine and enhance concepts from existing sorting methods discussed in Section \ref{sec:sota_mapping}. Our aim is to handle the increased data volume and complexity efficiently, while ensuring that regular sorting of Gaussian data during training does not negatively affect training time. We call our newly developed algorithm \textit{Parallel Linear Assignment Sorting (PLAS)}.

Our strategy involves an approximation approach to the highly complex assignment problem, inspired by the principles of the Fast Linear Assignment Sorting (FLAS) algorithm \cite{barthel2023}. We aim to find an optimal balance between computational efficiency and maintaining the accuracy of the spatial relationships among the Gaussian attributes. The parallel assignment process is visualized with random RGB colors in Figure \ref{fig:blockwise_assignment}: We initialize our grids by mapping the Gaussians to random positions to avoid getting stuck in a local minimum. Subsequently, we perform a low-pass filtering operation on the grid to construct an idealized target grid, and re-assign all elements to their best-matching positions regarding the smooth target. This process is repeated while decreasing the filter size, gradually sorting the grid.

For each re-assignment, the grid is divided into multiple blocks, which are processed independently. The block size $\beta$ is set as $\beta = \phi + 1$, where $\phi$ is the radius of the Gaussian blur applied to compute the target. We do not create blocks smaller than $\beta=16$, as tiny blocks inhibit efficient parallelization. To be able to sort across block borders and to cover all borders of the grid when not aligned at multiples of the current block size, we shift all blocks by a random $\Delta_y$ and $\Delta_x$ before starting re-assignment.

\begin{figure}[htbp]
  \centering
   \includegraphics[width=0.4\linewidth]{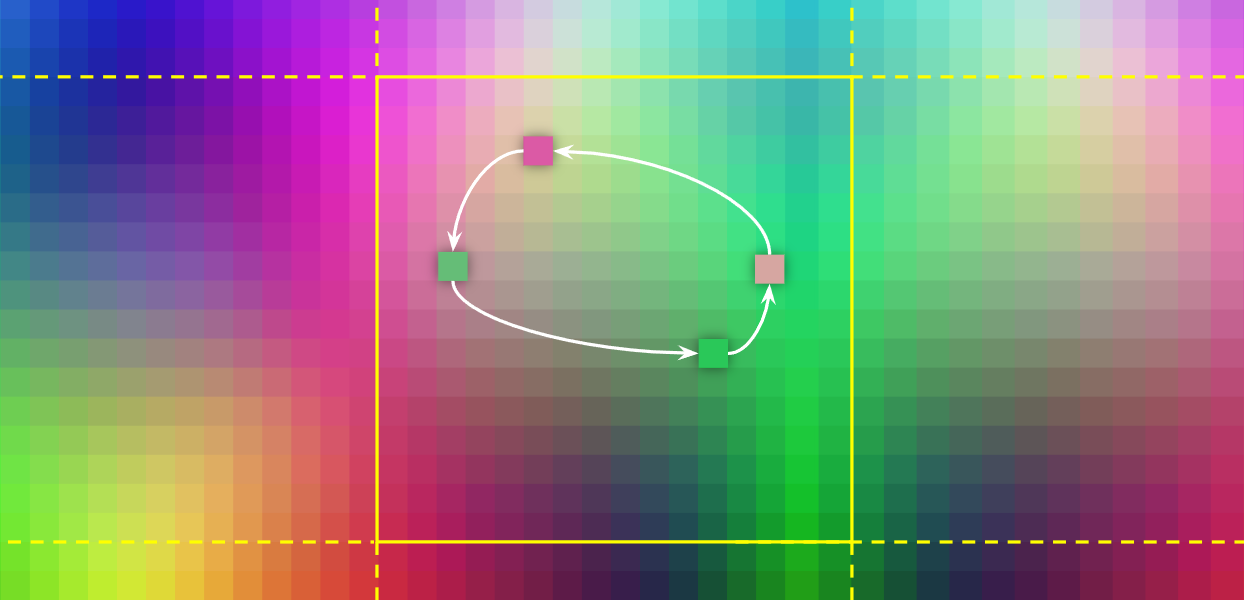}
   \vspace{-0.2cm}
   \caption{All values within each sub-block of the 2D grid are re-assigned in the same step. The elements from the block are split into randomly assigned, non-overlapping groups of 4. Out of the 24 possible permutations for each 4 elements, the positions that have the lowest total distance to the smoothed grid target are chosen.}
   \label{fig:swapping_pixels}
\end{figure}

After each block-wise linear assignment, we measure the mean L2 distance between sorted grid and target grid and compare it to its previous value. If it has improved by less than $0.01\%$, we start with a new random block position. When the assignment with the new block positions does not improve the distance further (using the same threshold), the radius is reduced by applying $\phi = 0.95 \phi$. We start with a radius $\phi = \frac{w}{2} - 1$, where $w$ is the total grid width, and we finish the work when the radius value would fall below 1, and the sorting has converged.

\noindent{\textbf{Blockwise Linear Assignment:}} As traditional sorting algorithms do not map well to GPU execution, we perform the assignment by swapping four elements at a time, as done in Self-Organizing Structured Maps \cite{Strong2013}. The groups of four are chosen from a random permutation of all elements from a block, using each element only once. The random grouping is important for pairing up elements with possible exchange targets from different areas within the block. All blocks are processed in parallel, and use the same permutation for the grouping. Finally, the new assignment for the groups of four employs a brute-force approach, iterating through the cost for all 24 possible permutations to find the optimal arrangement of elements. A single re-assignment is visualized in Figure \ref{fig:swapping_pixels}.

\noindent{\textbf{Organizing Gaussians into a 2D Grid:}} The presented algorithm is implemented highly parallel on the GPU, and will organize millions of Gaussians in a few seconds. In our application, we normalize all attributes of the Gaussians (coordinates, scale, etc.) independently to the range \([0, 1]\) and then scale them with individual weights before calculating the distance of the possible re-assignments. We use the activated values for all attributes (e.g. exponentially activated scale), as they would be seen by the renderer. At the start of the training, we sort once, and continue sorting with each densification step. As the number of Gaussians grows over time, so will the the size of the grid. We find the largest square grid size that can be completely filled with Gaussians for simplicity, and prune the Gaussians with the lowest opacity values that will not fit.

\subsection{Smoothness Regularization}
\label{subsec:smooth}
2D image compression methods work best if the content exhibits little noise and shows smooth structures that can be accurately predicted from neighboring values or represented by only a few transform coefficients. To further enforce such features in our 2D attribute grids, we employ a smoothness regularization on the sorted Gaussians during the training for scene optimization. Here, we exploit the fact that different Gaussian configurations can lead to similar visual quality while varying in local 2D grid smoothness.

For the regularization, we first create a blurred and thus smoother version of the 2D grid by applying a 2D Gaussian filter on all attribute channels. Deviations of the current Gaussian parameters at an original 2D grid position from those of the smoothed grid form the additional smoothness loss for training. For comparison, we select a Huber loss, as it is less sensitive to outliers than MSE. This has shown to improve the compression rate. 
We add the loss of the smoothness regularization to the loss of the 3D Gaussian Splatting algorithm using a weight $\lambda$.  As we propagate the gradient of the smoothness regularization through the Gaussian renderer during training, we specifically enforce the 3D Gaussian Splatting optimization process to prefer Gaussians that improve the quality of the rendering while also considering the respective local smoothness. This differentiates our method from compression methods that focus on reducing existing data in a pure post-processing step. Instead, we manipulate the data during creation to achieve high compression and high visual quality afterwards.

\subsection{Coordinate space contraction}
\label{sec:space_contraction}

We introduce an additional exponential activation for the coordinates $x$ used during training, allowing for more relative precision in coordinates in the center of the scene:

\begin{equation}
    x = \text{sign}(x_{\text{log}}) \times (\exp(|x_{\text{log}}|) - 1)
\end{equation}

Some NeRF implementations have used space contraction for unbounded scenes, notably Mip-NeRF 360 \cite{barron2021}. We have chosen to use this simpler logarithmic contraction function, which is easily invertible, and in design closer to the activation functions for scale and opacity used in 3DGS. It supports unbounded values, both negative and positive, as well as 0, all of which are required for the cartesian coordinates.

\subsection{Quantization \& Compression}
\label{sec:compression}
Once training has finished, we can use off-the-shelf image compression methods to store the data on disk. We store the un-activated Gaussian parameters, retaining the activation methods from 3DGS: sigmoid actiation for opacity and exponential activation for the scale, and applying our own logarithmic space contraction method (see Section \ref{sec:space_contraction}).

We clip the RGB values (DC values of the spherical harmonics) to the range \([-2, 4]\). Opacity values are clipped to the range \([-6, 12]\), non-DC SH features to \([-1, 1]\), rotations to \([-1, 2]\). These ranges are chosen to cover the 1st to 99th percentile of these values across all models from the 360 dataset. All but the scaling values are then normalized to the range \([0, 1]\).

We quantize the different attributes by rounding them to the closest value of a linear range of $q$ total values, with $q_{\text{coords}}=2^{14}$, 
$q_{\text{scale}}=q_{\text{opacity}}=q_{\text{rotation}}=2^6$ and $q_{\text{SH\_rest}}=2^5$. We compress the RGB grid with lossy JPEG~XL, as an 8-Bit image with a quality level of 100. All other attributes are stored as lossless JPEG XL. Our method is not restricted to this particular codec, but could also be used with other, existing 2D coding techniques.

\section{Evaluation}

In the following, we will perform several experiments to compare the quality and storage size to state-of-the-art 3D scene representation methods.
Our experiments are based off the official 3DGS implementation on GitHub\footnote{https://github.com/graphdeco-inria/gaussian-splatting}.
The parallel sorting is implemented in PyTorch and also uses the CUDA backend. An evaluation on the runtime performance of the sorting algorithm is given in the supplementary material. For our comparisons to prior methods, we select three real-world 3D scene reconstruction datasets: Mip-NeRF360, Tanks\&Temples and Deep Blending and a synthetic dataset: Synthetic-NeRF.

\begin{table*}
    \centering
    \begin{adjustbox}{width=12.2cm,center}
    \begin{tabular}{l|ccc|ccc|ccc|ccc}
        Dataset & \multicolumn{3}{c|}{Mip-NeRF360} & \multicolumn{3}{c|}{Tanks\&Temples}  & \multicolumn{3}{c|}{Deep Blending} & \multicolumn{3}{c}{Synthetic-NeRF}\\
        Method/Metric &  PSNR$\uparrow$ & LPIPS$\downarrow$ & Size$\downarrow$ &    PSNR$\uparrow$ & LPIPS$\downarrow$ & Size$\downarrow$ & PSNR$\uparrow$ & LPIPS$\downarrow$ & Size$\downarrow$ &  PSNR$\uparrow$ & LPIPS$\downarrow$ & Size\\
        \hline
        
        Plenoxels $\dagger$ & 23.08 & 0.463 & 2100 & 21.08 & 0.379 & 2300 & 23.06 & 0.510 & 2700 &  31.76 & - & -\\
        
        M-NeRF360 $\dagger$ & 27.69 & 0.237 & 8.6 & 22.22 & 0.257 & 8.6 & 29.40 & 0.245 & 8.6 & 33.09 & - & - \\
        
        INGP-Base $\dagger$ & 25.30 & 0.371 & 13 & 21.72 & 0.330 & 13 & 23.62 & 0.423 & 13 & 33.18 & - & -\\
        
        INGP-Big $\dagger$ & 25.59 & 0.331 & 48 & 21.92 & 0.305 & 48 & 24.96 & 0.390 & 48 & - & - & - \\
        
        VQ-TensoRF $\dagger$ & - & - & - & 28.20 & - & 3.3  & - & - & - & 32.86 & - & 3.6 \\
        
        \hline
        
        3DGS  &	27.55	&	0.222	&	785 &	25.54	&	0.201	&	454 &	30.07	&	0.248	&	699 &	33.88	&	0.031	&	71.6  \\
        3DGS w/o SH &	26.94	&	0.234	&	212 &	25.10	&	0.210	&	122 &	30.24	&	0.249	&	191 &	32.06	&	0.039	&	19.4   \\

        \hline
        \textbf{Ours} &	27.64	&	0.220	&	40.3 &	25.63	&	0.208	&	21.4  &	30.35	&	0.258	&	16.8  &	33.70	&	0.031	&	4.1 \\
        \textbf{Ours} w/o SH &	27.02	&	0.232	&	16.7  &	25.27	&	0.217	&	8.2   &	30.50	&	0.261	&	5.5   &	31.75	&	0.040	&	2.0\\

    \end{tabular}
    \end{adjustbox}
    \vspace{0.2cm}
    \caption{A comparison of our method to the default 3DGS with and without spherical harmonics (SH) and prior NeRF-based renderer. All sizes are in MB. Results with a $\dagger$ are directly copied from \cite{kerbl2023,Li2023}. Results for SSIM are included in the supplementary material.}
    \label{tab:experiments}

\end{table*}

\vspace{-1cm}
\noindent{\textbf{Quantitative results:}} In Table \ref{tab:experiments}, we compare our method to the default 3D Gaussian Splatting algorithm and prior NeRF-based 3D reconstruction methods, reporting the standard PSNR and L-PIPS metrics. Variants marked \textit{w/o SH} are using only the DC part of the spherical harmonics to provide color on the Gaussian Splats, and deactivating any view-dependent effects (higher-level spherical harmonics). %

We demonstrate that we reduce the average storage size by a factor of 17x to 42x (depending on the dataset), compared with 3DGS \cite{kerbl2023}, without sacrificing visual quality. The highest reduction of 41.6x is observed on the Deep Blending dataset. Training our method with spherical harmonics deactivated yields a reduction factor of 127x over vanilla 3DGS while improving in PSNR.

For some datasets, compared to Mip-NeRF360 \cite{barron2021} and VQ-TensoRF \cite{Li2023}, we achieve a slightly higher PSNR and slightly lower L-PIPS, but at the same time allow for real time rendering and fast training, given the efficient rasterization algorithm of 3DGS \cite{kerbl2023}.
Additionally, while Mip-NeRF360 trains one scene for several hours, our method only uses 10 to 30 minutes. 
Most notably, we achieve the same training time as 3DGS, even though our method brings some computational overhead compared to the default 3DGS, arising from the periodic sorting of Gaussians and the necessity to blur grids for computing the neighbor loss. This is due to our parameter configuration enforcing the optimization process to create a considerably smaller number of Gaussians. For example, for the \textit{Truck} dataset, our method creates 1.55M Gaussians, whereas the default 3DGS creates 2.58M. %
For the same reason of the reduced number of Gaussians, our scenes render much faster in the 3DGS viewer than the original models. The vanilla \textit{Truck} with 2.58M Gaussians renders at 385 fps with the default settings on the same GPU, while our scene with 1.55M Gaussians renders at 515 fps with better visual quality.

\begin{figure*}[htbp]
  \centering
   \includegraphics[width=0.95\linewidth]{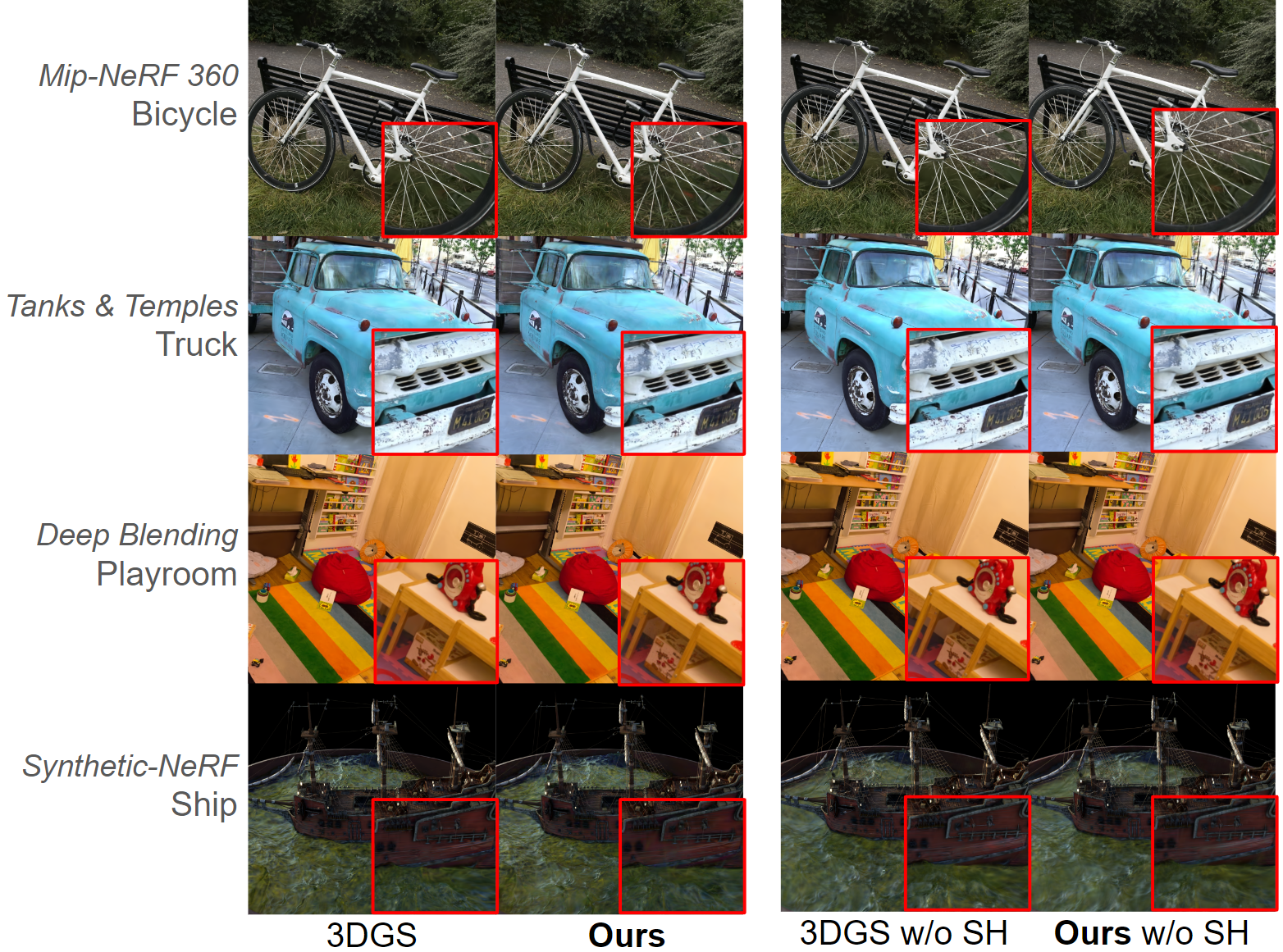}
   \caption{A qualitative comparison across example scenes from all four datasets.}
   \vspace{-0.3cm}
   \label{fig:quality_compare}
\end{figure*}

\noindent{\textbf{Qualitative results:}} The results of the metrics suggest that the rendering quality of our method is very similar to the vanilla 3DGS. This is also demonstrated in figure \ref{fig:quality_compare}, where we directly compare the renderings from different datasets side by side. Across all examples we observe very similar quality compared to vanilla 3DGS. This underlines that our method is able to maintain a high rendering quality, while only using a fraction of the storage size. 

\begin{figure*}[htbp]
  \centering
   \includegraphics[width=0.9\linewidth]{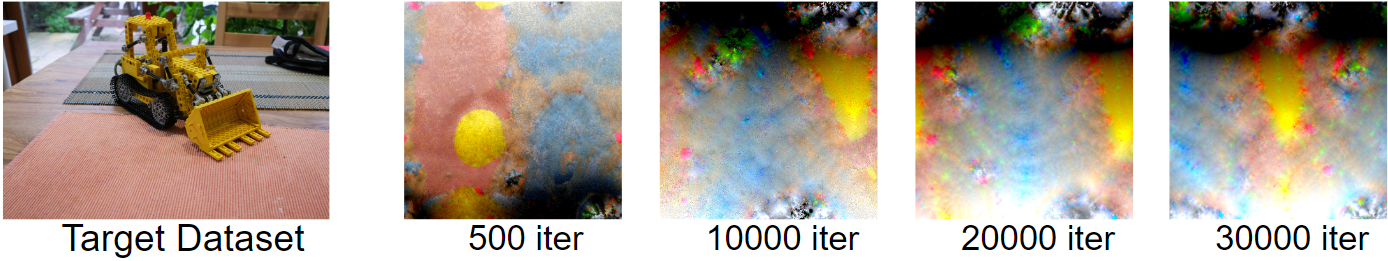}
   \vspace{-0.2cm}
   \caption{Resulting 2D attribute grids for the color. Until iteration 15000, the number of Gaussians grow from about 100k up to 1.5 million.}
   \label{fig:rgb_grids}
\end{figure*}

Figure \ref{fig:rgb_grids} displays examples of the sorted 2D grids during the training process. Specifically, we show the 2D RGB color grids for thekitchen dataset. We observe that the colors of the scenes organize themselves into clusters of similar colors. The grids, however, do not appear to be more organized with more iterations. This has two reasons. Firstly, the number of Gaussians grows substantially during the training, posing an increasingly more difficult sorting task. And secondly, as our method has to sort all attributes at once using the same permutation, isolated features, such as the color cannot be organized perfectly.

\subsection{Smoothness Regularization}
The magnitude of the smoothness regularization $\lambda$ decides the size of the gradient for local smoothness in the 2D attribute grids in relation to the size of the gradient of the Gaussian training algorithm. To find a good balance between both optimization targets, we tested a series of parameters for $\lambda$, see table \ref{tab:lambda_ablation}. As expected, lower magnitudes for $\lambda$ result in higher rendering quality, but also larger storage size.

\vspace{-0.5cm}
\begin{table}[htbp]
    \begin{minipage}{.3\linewidth}
    \centering
    \begin{tabular}{|c|c|c|}
        \hline
        $\lambda$ & PSNR & Size in MB \\
        \hline
        0.01 & 24.39 & 18.02 \\
        0.05 & \textbf{24.48} & 18.35 \\
        0.1  & 24.18 & 17.50 \\
        0.5  & 24.40 & 14.62 \\
        1.0  & 24.24 & 12.43 \\
        1.5  & 24.02 & \textbf{10.73} \\
        \hline
    \end{tabular}
    \vspace{0.2cm}
    \caption{Testing different magnitudes for the neighbor loss $\lambda$. }
    \label{tab:lambda_ablation}
    \end{minipage}
    \begin{minipage}{.7\linewidth}
    \centering
        
    \begin{tabular}{|l|rr|rr|}
    \hline
    & \multicolumn{2}{c|}{w/ SH} & \multicolumn{2}{c|}{w/o SH}  \\
    Method &	PSNR	&	Size	&	PSNR	&	 Size	    \\
    \hline
    Ours        &	25.37	&	34.3	&	24.90	&	17.3	\\
    \hline
    3DGS Our Compression  &	21.98	&	71.7	&	20.37	&	44.5\\
    \hline
    3DGS .ply	      &	25.44	&	624.7	&	24.89	&	174.0	\\
    3DGS .ply.zip	      &	25.44	&	548.2	&	24.89	&	130.8	\\
    Blog Post \cite{blog_post} & 25.05 & 41.5   &   -      &    -     \\
    \hline
    \end{tabular}
    \vspace{0.2cm}
    \caption{Comparison of our method to other 3DGS compression approaches, calculated using the Truck scene from Tanks\&Temples. Size is in MB.}
    \label{tab:ablation}

    \end{minipage}
\end{table}

\vspace{-1cm}
\subsection{Synergy of Sorting and Compression}
\label{sec:synergy}

To validate the effectiveness of the sorting and smoothing during the training, we measure the storage size and PSNR of a model that is only compressed after the vanilla 3DGS training. The result is shown in table \ref{tab:ablation} \textit{3DGS Our Compression}. We observe a much lower PSNR compared to our result with sorting and smoothing activated. This underlines that the local smoothness, caused by sorting and smoothing, plays a significant role for the success of our approach.

To put our approach into relation with available compression methods we perform two further comparisons in Table \ref{tab:ablation}. Firstly, we simply compress the 3DGS .ply file with a zip compression. And secondly, we compare the results of the method developed in the blog post series \textit{Making Gaussian Splats smaller} \cite{blog_post}. As a result, we observe that our method outperforms the .zip compression substantially, reducing the storage size by a factor of 16x. For the comparison with \textit{Making Gaussian Splats smaller}, the results are closer, nevertheless, our method achieves both higher PSNR and lower storage size.

It may be possible to encode the Gaussian Splats with conventional point cloud compression methods. The popular DRACO algorithm \cite{draco2023} supports encoding additional attributes on top of position and color in theory, but its implementations do not \cite{dracoIssue757}. Thus, a direct comparison is unfortunately not possible. Our method's strength lies in adapting the representation during training to be well compressible (see Section \ref{sec:synergy}). Using any post-training compression method, like DRACO, cannot make use of these gains.

\begin{table}[htbp]
\centering
\begin{tabular}{|r|c|c|c|c|}
\hline
Name & PSNR $\uparrow$ & SSIM $\uparrow$ & LPIPS $\downarrow$ & Size (MB) \\
\hline
PLY & 25.41 & 0.880 & 0.152 & 104.89 \\
NPZ & 25.41 & 0.880 & 0.152 & 76.10 \\
JPEG XL lossless & 25.41 & 0.880 & 0.152 & 66.05 \\
PNG 16 & 24.08 & 0.873 & 0.157 & 37.00 \\
EXR & 25.36 & 0.878 & 0.154 & 30.56 \\
\textbf{JPEG XL quantized} & 25.14 & 0.870 & 0.161 & \textbf{11.86} \\
\hline
\end{tabular}
\vspace{0.2cm}
\caption{The \textit{Truck} scene, trained with our method, compressed in different formats. Marked in bold is the compression method used in our experiments for all datasets.}
\label{tab:compression_exps}
\end{table}

\vspace{-1cm}
\subsection{Choice of compression format}

Table \ref{tab:compression_exps} contrasts different compression methods on the \textit{Truck} scene, without using spherical harmonics. The scene was trained with our parameters and sorted into 2D grids. Then we sampled different image and array encoding formats, measuring rendering quality and file size.%
Of the lossless formats, JPEG XL provides the smallest files. Notably, the PLY file when trained with our method is of smaller size and of higher quality than training with the default 3DGS parameters (which yields 24.90~PSNR at 174~MB). \textit{NPZ} is using NumPy's \texttt{save\_compressed} method to directly store the tensors. \textit{JXL~ll} stores all tensors as lossless JPEG~XL images.

By introducing lossy compression, the file size can be further decreased, with different tradeoffs between rendering quality and storage space used. PNG~16 truncates the value below the 1st and above the 99th percentile (over all 360 datasets), then normalizes the data to the \([0, 1]\) range and stores them as 32-Bit PNGs. \textit{EXR} stores the values as 32-Bit OpenEXR files with zip compression. We have found using JPEG XL with different levels of quantization provides the best tradeoff. Even though we are not using spherical harmonics, our chosen compression method is providing close to the same quality as 3DGS with spherical harmonics (25.44~PSNR at 615~MB), at a compression rate of 52x.

\subsection{Features to 2D Grid}

\begin{figure}[htbp]
  \centering
   \includegraphics[width=0.9\linewidth]{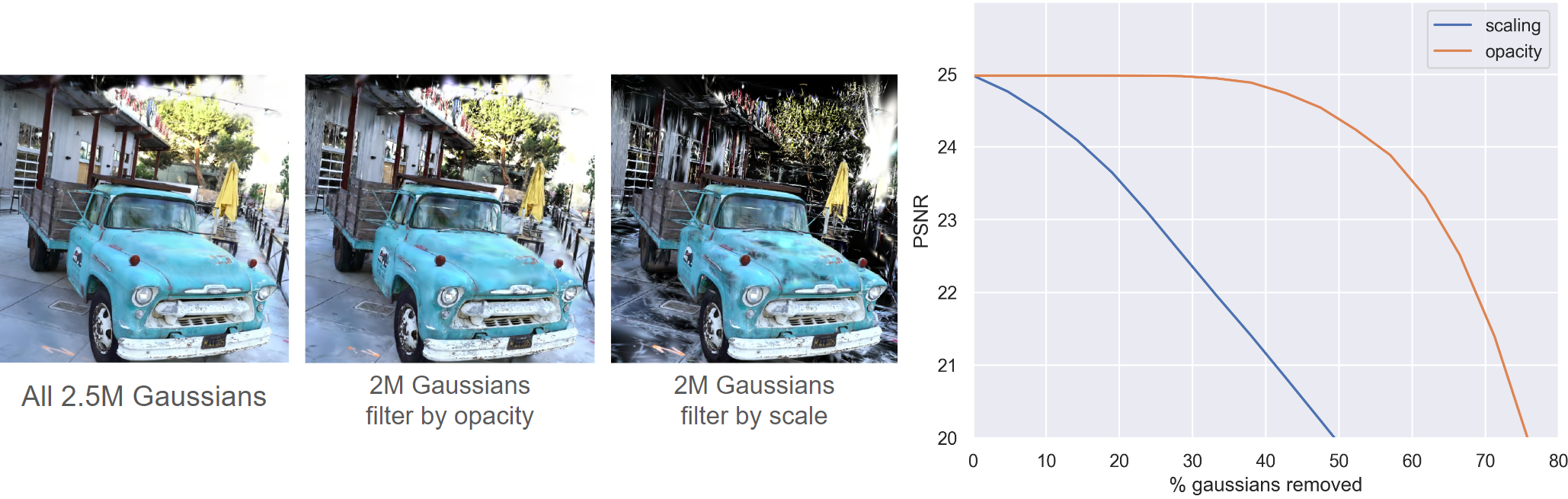}
   \vspace{-0.2cm}
   \caption{Removing the first \%  of Gaussians, ordered by scaling and opacity from the lowest to the highest.}
   \label{fig:remove_gaussians}
\end{figure}

To convert the millions of Gaussians into a 2D grid, we reshape the list of features into a square grid. As the number of Gaussians grows over time from the densification process, so will the the size of the grid. For simplicity, we find the largest square grid size that can be completely filled with Gaussians, and discard the Gaussians with the lowest opacity values that will not fit. Losing up to one row and one column of Gaussians does not meaningfully alter the scene: In figure \ref{fig:remove_gaussians}, we show that the PSNR of a rendered scene is not affected when removing up to 30\% of Gaussians with the lowest opacity. In contrast, removing Gaussians with small scaling, affects the quality of the scene considerably more.

\vspace{-0.5cm}

\vspace{-1.cm}
\section{Conclusion}
\label{sec:conclusion}

We have presented an optimized representation of 3D Gaussian Splatting, targeting efficient compression and storage. This is achieved by a novel, highly parallelized sorting strategy that smoothly arranges the high dimensional, initially unordered Gaussian parameter sets in a 2D grid, which can be encoded by standard image coding techniques like JPEG~XL. In addition, an optimized training scheme for the 3D Gaussians with a new smoothness loss leads to splat configurations with even higher smoothness in the 2D grid, while still preserving the original accuracy of the 3D scene. In spite of the proposed extensions, training time only negligibly changes  compared to the original 3DGS approach. With the proposed solution, we can reduce the data down to 2\% of the original size at the same visual quality, enabling the efficient handling of large 3DGS scenes in practical applications.

In future work, we aim to further enhance compression efficiency. Besides an adapted weighting of the individual parameters of the smoothness loss, better and less correlated representations for rotations, shape and spherical harmonics will be investigated. To achieve even better results, it would be interesting to perform the quantization during training, which is currently happening only once before the compression. The main focus, however, will be on the extension to 4D scenes with consideration of temporal dependencies, supporting the strength of 3D Gaussians in the representation of dynamic scenes.

\section*{Acknowledgements}

This work has partly been funded, by the German Federal Ministry for Economic Affairs and Climate Action (ToHyVe, grant no.~01MT22002A) and the German Research Foundation (3DIL, grant no.~502864329).

\bibliographystyle{splncs04}
\bibliography{main}

\newpage
\appendix

\title{Appendix} 
\author{}
\institute{}
\maketitle

\setcounter{section}{0}
\renewcommand{\thesection}{\Alph{section}}
\setcounter{subsection}{0}
\renewcommand{\thesubsection}{\Alph{section}.\arabic{subsection}}
\setcounter{subsubsection}{0} %

\section{Results for SSIM}

In addition to the results from table \ref{tab:experiments}, we also measure the structural similarity index measure (SSIM) in table \ref{tab:ssim}. These results correlate with the PSNR results from table \ref{tab:experiments}.

\begin{table*}
    \centering
    \begin{tabular}{l|cc|cc|cc|cc}
        Dataset & \multicolumn{2}{c|}{Mip-NeRF360} & \multicolumn{2}{c|}{Tanks\&Temples}  & \multicolumn{2}{c}{Deep Blending} & \multicolumn{2}{c}{Synthetic-NeRF}\\
        Method/Metric & SSIM$\uparrow$ & Size (MB) &  SSIM$\uparrow$ & Size (MB) &  SSIM$\uparrow$ & Size (MB) & SSIM$\uparrow$ & Size (MB) \\
        \hline
        
        Plenoxels $\dagger$ & 0.626 & 2100 & 0.719 & 2300 & 0.795 & 2700 & - & -\\
        
        M-NeRF360 $\dagger$ & 0.792 & 8.6 & 0.759 & 8.6 & 0.901 & 8.6 & - & - \\
        
        INGP-Base $\dagger$ & 0.671 & 13 & 0.723 & 13 & 0.797 & 13 & - & -\\
        
        INGP-Big $\dagger$ & 0.699 & 48& 0.745 & 48 & 0.817 & 48 & - & - \\
        
        VQ-TensoRF $\dagger$ & - & - &  0.913 & 3.3 & - & - & 0.960 & 3.6 \\
        \hline
        3DGS &	0.814	&	785 &	0.866	&	454 &	0.907	&	699 &	0.970	&	71.6 \\
        3DGS w/o SH &	0.803	&	212 &	0.859	&	122 &	0.908	&	191 &	0.962	&	19.4\\
        \hline
        Ours &	0.814	&	40.3 &	0.864	&	21.4 &	0.909	&	16.8 &	0.969	&	4.1 \\
        Ours w/o SH &	0.803	&	16.7 &	0.857	&	8.2 &	0.908	&	5.5 &	0.961	&	2.0 \\

    \end{tabular}
    \vspace{0.2cm}
    \caption{A comparison over structural similarity index measure (SSIM) between our method, the default 3DGS model and prior NeRF-based methods. Results with a $\dagger$ are directly copied from \cite{kerbl2023,Li2023}. Our method achieves close to the same SSIM as vanilla 3DGS, while reducing size with a factor of 17x to 42x, depending on the dataset.}
    \label{tab:ssim}
\end{table*}

\section{Parameter Selection}

The 3D Gaussian Splatting (3DGS) algorithm involves many different training parameters. In the following, we will give a brief overview on these parameters and highlight those we changed for our new training algorithm.

\subsection{Densification}
To minimize the sorting time during training, we make minimal changes to the default parameters of 3DGS, i.e.~we reduce the number of Gaussians that are created during optimization. To achieve this, we modify the following five parameters: 

\begin{itemize}
    \item \textit{Densification interval} $\in \mathbb{N}$: Determines how often the densification process executed. A low value corresponds to frequent densification, which results in a large amount of Gaussians.
    \item \textit{Densify grad threshold} $\in \mathbb{R}$: All Gaussians with a larger accumulated gradient for the xyz position are split and cloned. A low threshold results in more Gaussians generated during training.
    \item \textit{Densify min opacity} $\in \mathbb{R}$: This threshold filters all Gaussians with smaller opacity during densification. A high value leads to fewer Gaussians. 
    \item \textit{Opacity reset interval} $\in \mathbb{N}$: Every $x$ steps, the opacity of all Gaussians is set to 0.01.
    \item \textit{Percent dense} $\in \mathbb{R}$: During the densification process, large Gaussians are split into two smaller copies, with 80\% of the original size each, while small Gaussians are cloned identically. The percent dense parameter specifies the threshold at which a Gaussian is classified as small or large. The value, which  is set between 0 and 1, is multiplied by the extend of the scene and used as comparison.
\end{itemize}

\begin{table}[htbp]

\begin{minipage}[t]{.55\linewidth}
    \centering
    \begin{tabular}{|l|c|c|}
        \hline
        Parameter / Method & 3DGS & Ours \\
        \hline
        \textbf{3DGS} &  &  \\
        Densification interval & 100 & 1000 \\

        Densify grad threshold & $2 \times 10^{-4}$ & $7 \times 10^{-5}$ \\
        Densify min opacity & 0.005 & 0.1\\
        Opacity reset interval & 3000 & $\infty$ \\
        Percent dense & 0.01 & 0.1 \\
        \hline
        \textbf{Smoothness Reg.} &  & \\
        Kernel size & - & 5 \\
        Sigma &  - & 3 \\
        Overall multiplier $\lambda$ & - & 1.0 \\
        \textbf{Weights} &  &  \\
        Position & - & 0.0 \\
        Color (SH DC) & - & 0.0 \\
        Opacity & - & 0.09 \\
        Scaling & - & 0.0 \\
        Rotation & - & 0.91 \\
        Sph. Harmonics (rest) & - & 0.0 \\
        \hline
        \textbf{Sorting weights} &  &  \\
        Position & - & 1.0 \\
        Color (SH DC) & - & 1.0 \\
        Opacity & - & 0.0 \\
        Scaling & - & 1.0 \\
        Rotation & - & 0.0 \\
        Sph. Harmonics (rest) & - & 0.0 \\
        \hline
    \end{tabular}
    \vspace{0.5cm}
    \caption{A detailed list of our training parameters compared to the default 3DGS training. An opacity reset interval of $\infty$ denotes that we deactivate the opacity reset. }
    \label{tab:params}
\end{minipage}\hspace{.05\linewidth}
\begin{minipage}[t]{.4\linewidth}
    \centering
    \includegraphics[width=1.0\linewidth]{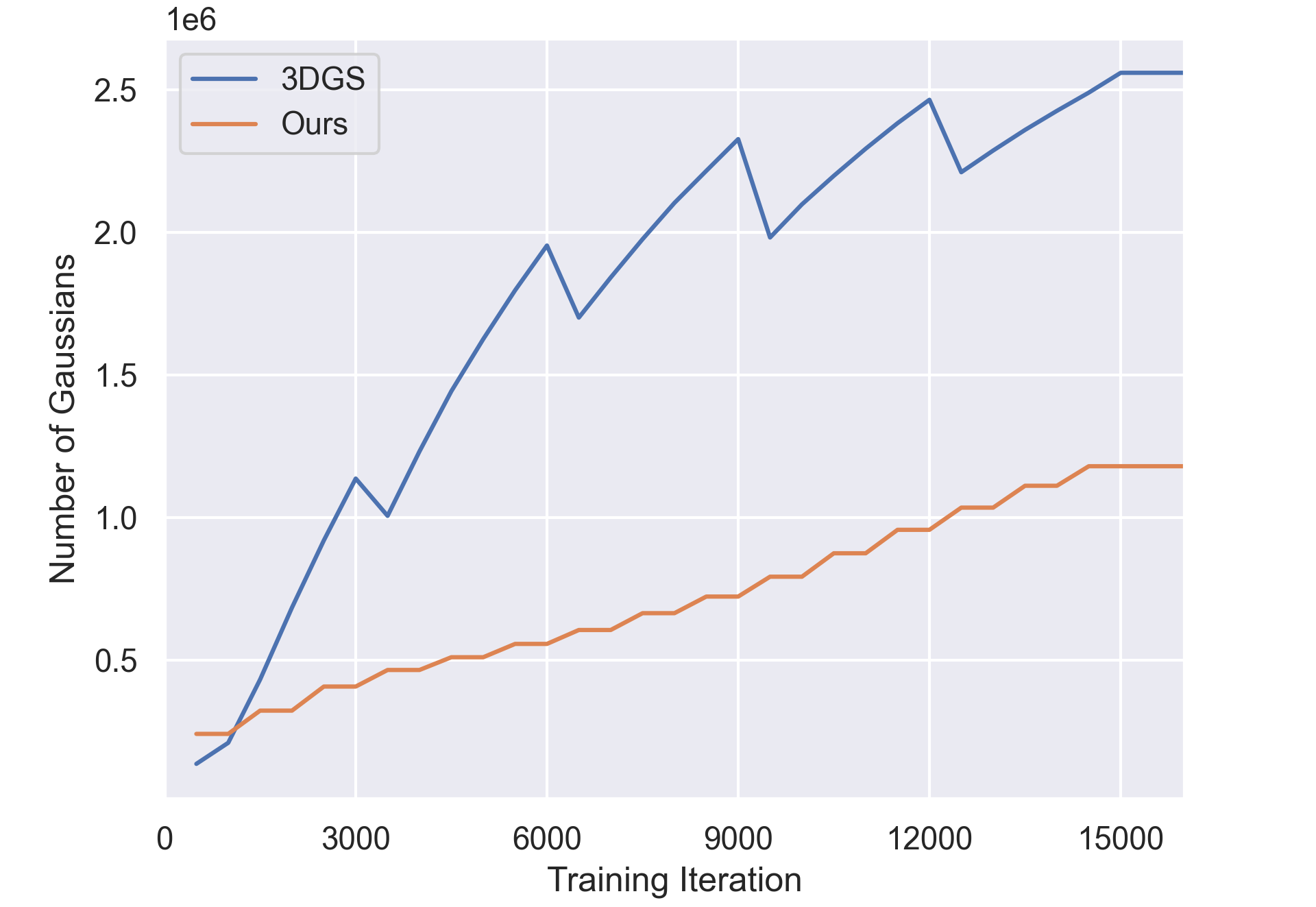}
    \vspace{0.5cm}
    \caption{By changing 3DGS default parameters that control the densification step, we observe a different number of Gaussians during the training. Our method creates less Gaussian splats and does not apply regular pruning, which happens every 3k steps with the default parameters.}
    \label{fig:num}
\end{minipage}
\end{table}

\subsection{Smoothness Regulation}
The smoothness regularization first blurs the 2D attribute grids and then computes an error term with a Huber loss. This results in new parameters to tune:

\begin{itemize}
    \item \textit{Kernel size} $\in \mathbb{N}$: The size of the blur filter. A higher value leads to smoother target images.
    \item \textit{Sigma} $\in \mathbb{R}$: The standard deviation of the Gaussian blur.
    \item \textit{Overall multiplier} $\lambda \in \mathbb{R}$: A multiplier scaling the loss of the Smoothness Regularization before adding it to the loss of 3DGS.
    \item \textit{Separate multiplier} $\in \mathbb{R}$: A loss scalar for each of the five 2D attribute grids (position, color, opacity, scaling and rotation). 
\end{itemize}

\subsection{Sorting}
The sorting algorithm that takes place after every densification can be adjusted with separate multipliers for each of the 3DGS attributes: position, spherical harmonics (DC and rest), opacity, scaling and rotation. In Table \ref{tab:params}, we contrast the vanilla 3DGS parameters with the ones we chose.

\subsection{Hyperparameter sensitivity}
We found our current sets of parameters through an iterative search, optimizing with the goal to stay close to vanilla 3DGS quality, while minimizing file size.  
Our guideline for the found parameters is the following: position, color and scale have a large effect on the PSNR, thus they were selected as the sorting keys. Opacity and rotation have a much smaller effect. Therefore, they are smoothed in their neighborhoods instead (with neighbors defined by position/color/scale), to be compressible.

We applied the same parameter sets for the evaluation across all datasets, including synthetic ones, demonstrating their generalizability. Superior configurations might be uncovered through an exhaustive parameter sweep. We think performing this computationally expensive work is best done in future, after choosing a more compact representation than the high-dimensional spherical harmonics attributes, for view-dependent effects.

\newpage
\section{Sorting performance}

In this section, we provide additional measurements of the performance of our novel sorting algorithm.
\begin{figure}[htbp]

\begin{minipage}[t]{.48\linewidth}
      \centering
       \includegraphics[width=1.0\linewidth]{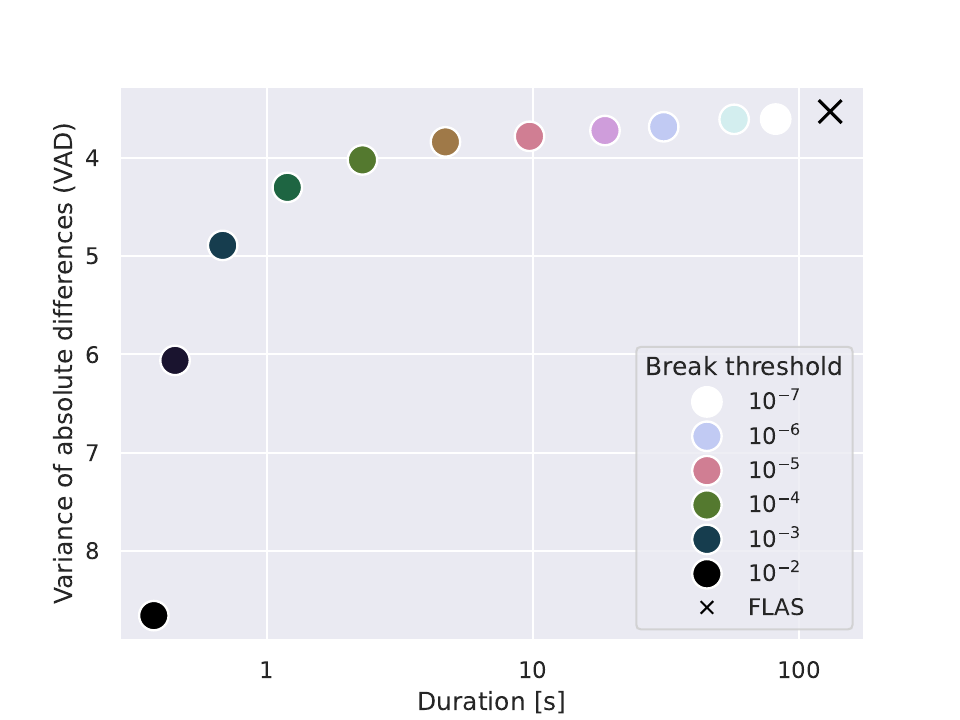}
       \caption{Behavior of the sorting algorithm under different break thresholds. Decreasing the relative L2 threshold trades off additional sorting quality for longer runtime. These values were measured with an \textit{NVidia RTX 4090} on a random 512x512x3 grid. In all our training experiments, we fixed this parameter to $10^{-4}$.}
       \label{fig:vad_vs_ib}
\end{minipage}\hspace{.04\linewidth}
\begin{minipage}[t]{.48\linewidth}
      \centering
       \includegraphics[width=1.0\linewidth]{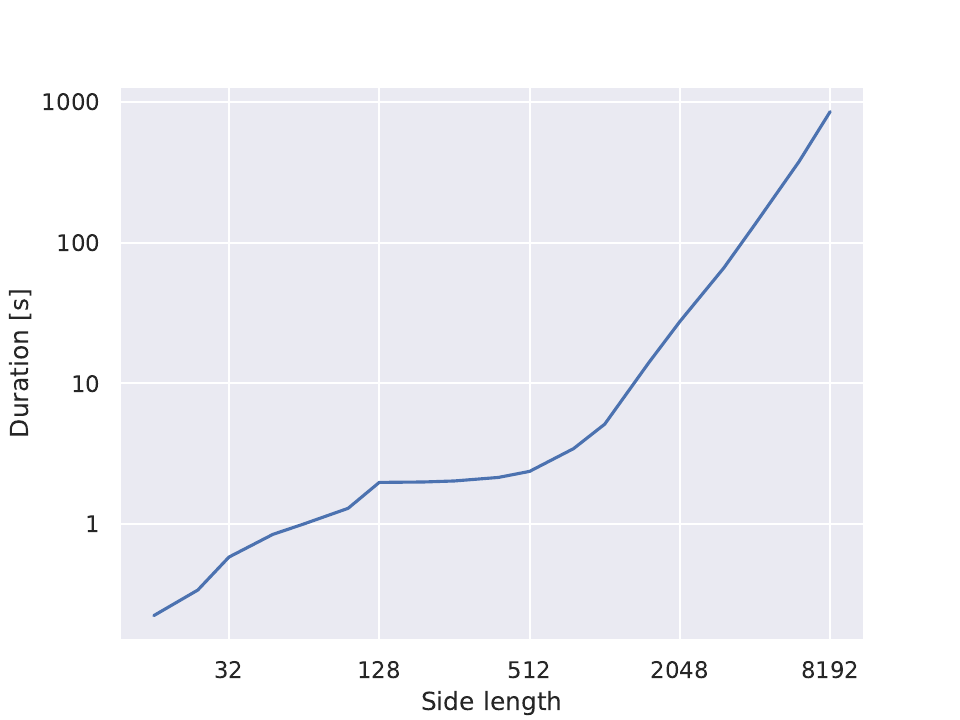}
       \caption{Runtime of the sorting algorithm over the side length of the 2D grid. Here measured using 3 layers for the grids, while there may be up to 14 when sorting Gaussians.}
       \label{fig:blocky_perf}
\end{minipage}
\end{figure}

To demonstrate the effectiveness of our parallel sorting algorithm, we compare it to the state of the art in quality, FLAS \cite{barthel2023}. An implementation of Self-Sorting Maps \cite{Strong2013} was not available to us. We measure runtime and, to estimate the quality of the sort, the variance of the absolute differences (VAD) between all neighboring values, over all channels.
Sorting a random 512x512x3 grid on an \textit{AMD Ryzen Threadripper PRO 5955WX} CPU with FLAS takes 131s and achieves a VAD of 3.53. Our algorithm running on an \textit{Nvidia RTX 4090} finishes in 5.7 seconds after 8015 reorders with a VAD of 4.02 (the shuffled data has a VAD of 3607.45). We therefore attain a sorting quality comparable to that of FLAS but with a significantly reduced runtime, thanks to our algorithm's high degree of parallelism.

The parameter that has the largest influence on the grid sorting performance is the  threshold, which decides whether the relative reduction in L2 distance between two iterations should stop sorting with the current configuration, and potentially continue with the next permutation or next lower level. Decreasing it leads to a higher sorting quality, but the increase in iterations takes additional runtime. In Figure \ref{fig:vad_vs_ib}, we have plotted the sorting quality over the total runtime.

In Figure \ref{fig:blocky_perf}, we show the runtime performance of the sorting algorithm depending on the input grid size. With the chosen parameters, sorting will usually take less than 10 seconds during training. The highest number of Gaussians of any of our models over all datasets is the \textit{Garden} scene with 4.37M Gaussians, which requires a grid of a side length of 2091. Thus, even the largest of scenes in the used datasets can be sorted in well below a minute.

\begin{figure*}[htbp]
  \centering
   \includegraphics[width=1.0\linewidth]{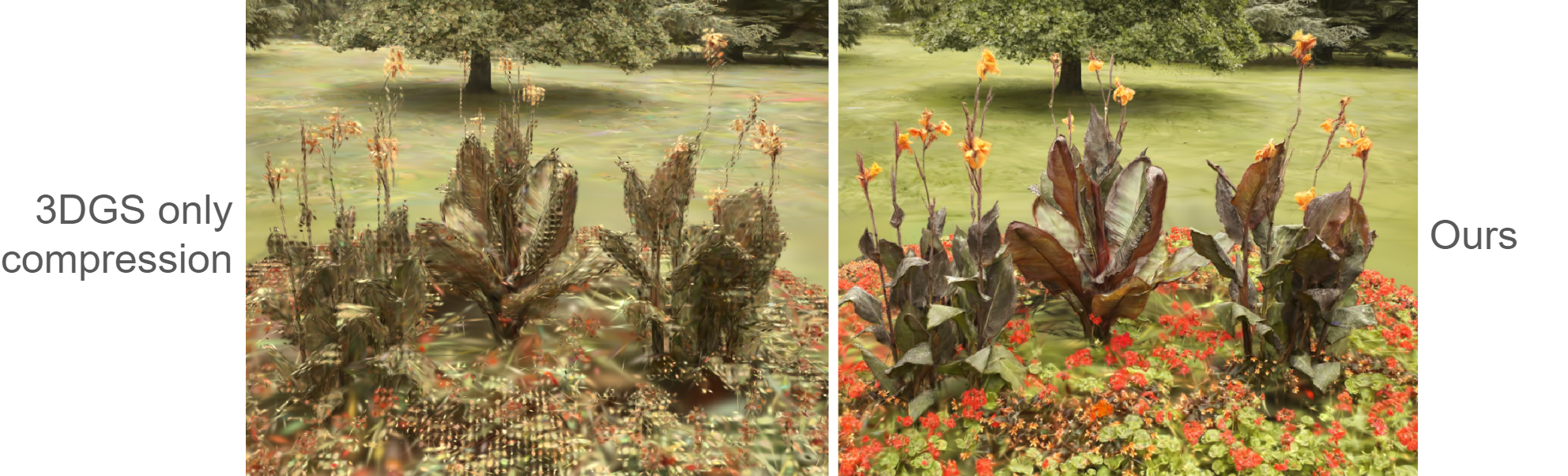}
   \vspace{-0.2cm}
   \caption{A comparison between the rendering quality when only applying our compression method to the vanilla 3DGS model without smoothing.}
   \label{fig:only_compress}
\end{figure*}

Table \ref{tab:ablation} suggests that the visual quality is drastically reduced when only applying sorting and compression after training a vanilla 3D Gaussian Splatting model. This can also be observed visually in Figure \ref{fig:only_compress}. This underlines that the smoothing regularization, which influences the 3DGS training, plays a significant role for compressing the attributes efficiently. 

\pagebreak

\end{document}